\documentclass{article}

\PassOptionsToPackage{numbers,square,sort&compress}{natbib}
\usepackage[preprint]{neurips_2025}

\usepackage[T1]{fontenc}    
\usepackage{url}            
\usepackage{booktabs}       
\usepackage{amsfonts}       
\usepackage{nicefrac}       
\usepackage{microtype}      
\usepackage{xcolor}         
\usepackage[table]{xcolor}
\usepackage{amsmath}
\usepackage{algorithm}
\usepackage{natbib}
\usepackage{algorithmicx}
\usepackage{algpseudocode} 
\usepackage{graphicx}
\usepackage{float} 
\usepackage{caption} 
\usepackage{multirow}
\usepackage{tabularx}  
\usepackage{booktabs}  
\usepackage{listings}
\usepackage{makecell}  
\usepackage{array}     
\usepackage{wrapfig}
\usepackage{enumitem} 
\title{\includegraphics[height=0.55cm]{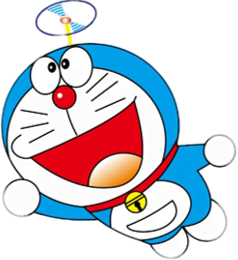}DORAEMON: Decentralized Ontology-aware Reliable Agent with Enhanced Memory Oriented Navigation}

\newcommand{\xin}[1]{\textcolor[rgb]{0.2,0.3,0.9}{#1}}

\author{%
  ~{Tianjun Gu}\textsuperscript{1} 
  \And
  ~{Linfeng Li}\textsuperscript{1} 
  \And
  ~{Xuhong Wang}\textsuperscript{2} 
  \And
  ~{Chenghua Gong}\textsuperscript{1} 
  \And
  ~{Jingyu Gong}\textsuperscript{1} 
  \And
  ~{Zhizhong Zhang}\textsuperscript{1} 
  \And
  ~{Yuan Xie}\textsuperscript{1,3} 
  \And
  ~{Lizhuang Ma}\textsuperscript{1} 
  \And
  ~{Xin Tan}\textsuperscript{1,2}
}

\begin{document}
\maketitle
\vspace{-1.5\baselineskip}
\begin{center} 
  \textsuperscript{1}East China Normal University,  
  \textsuperscript{2}Shanghai AI Lab,  
  \textsuperscript{3}Shanghai Innovation Institute
\end{center}

\begin{abstract}
Adaptive navigation in unfamiliar environments is crucial for household service robots but remains challenging due to the need for both low-level path planning and high-level scene understanding. While recent vision-language model (VLM) based zero-shot approaches reduce dependence on prior maps and scene-specific training data, they face significant limitations: spatiotemporal discontinuity from discrete observations, unstructured memory representations, and insufficient task understanding leading to navigation failures. We propose DORAEMON (Decentralized Ontology-aware Reliable Agent with Enhanced Memory Oriented Navigation), a novel cognitive-inspired, zero-shot, end-to-end framework consisting of Ventral and Dorsal Streams that mimics human navigation capabilities. The Dorsal Stream implements the Hierarchical Semantic-Spatial Fusion and Topology Map to handle spatiotemporal discontinuities, while the Ventral Stream combines CoDe-VLM and Exec-VLM to improve decision-making. Our approach also develops Nav-Ensurance to ensure navigation safety and efficiency. We evaluate DORAEMON on the HM3Dv1, HM3Dv2, MP3D, where it achieves state-of-the-art performance on both SR and SPL metrics, significantly outperforming existing methods. We also introduce a new evaluation metric (AORI) to assess navigation intelligence better. Comprehensive experiments demonstrate DORAEMON's effectiveness in zero-shot and end-to-end navigation without requiring prior map building or pre-training. Our code is available at \xin{\url{https://grady10086.github.io/DORAEMON/}}.

\end{abstract}

\section{Introduction}

\begin{wrapfigure}{r}{0.5\textwidth}
  \vspace{-48pt} 
  \centering
  \includegraphics[width=0.45\textwidth]{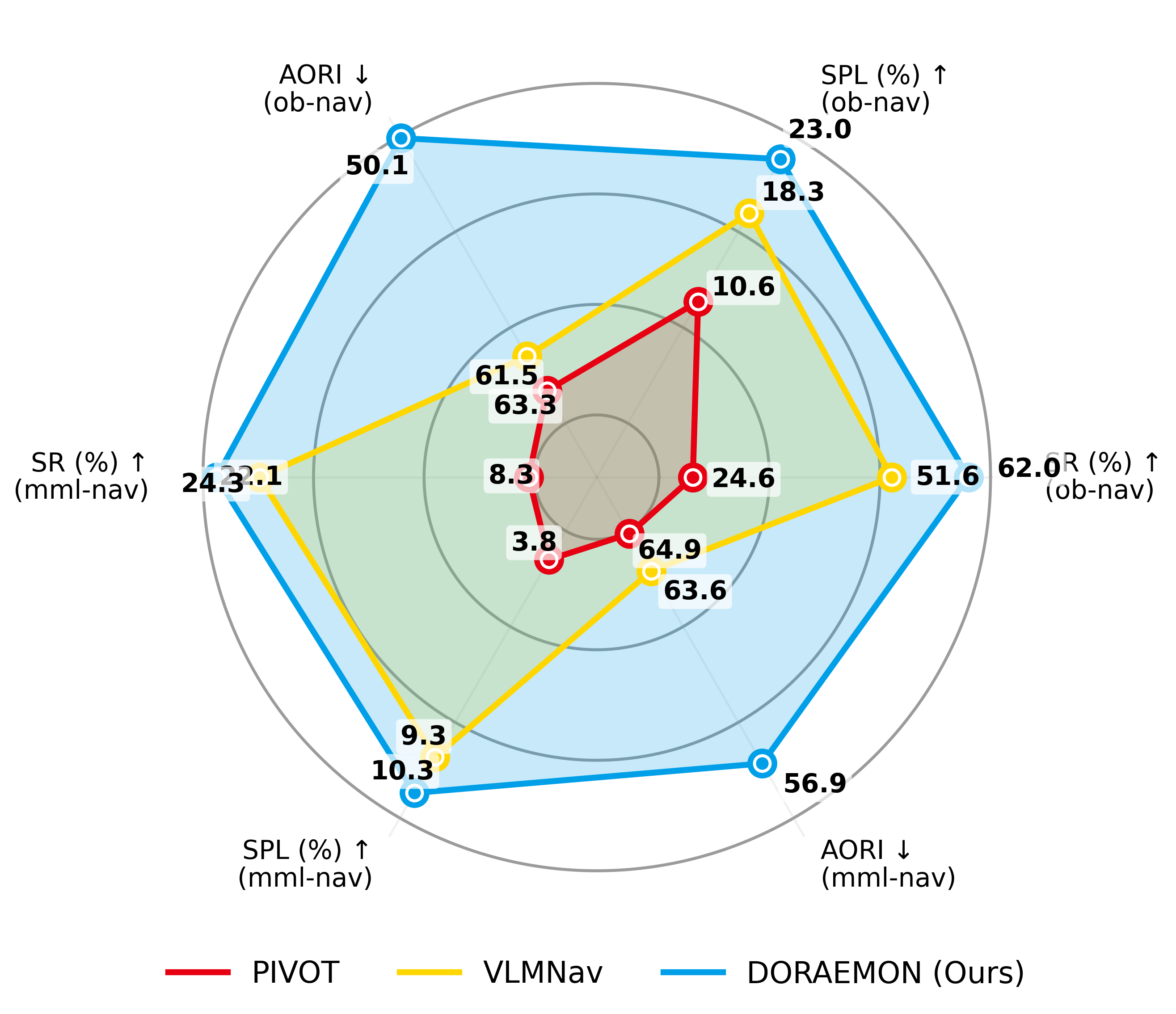}
  \caption{Performance comparison of end-to-end methods in object navigation(ob-nav) and multi-modal lifelong navigation(mml).}
  \label{fig:radar_comparison}
  \vspace{-19pt} 
\end{wrapfigure}

Adaptive navigation in complex and unseen environments \cite{DBLP:journals/corr/abs-2006-13171}  is a key capability for household service robots. This task requires robots to move from a random starting point to the location of a target object without any prior knowledge of the environment. For humans, navigation appears almost trivial, however,navigation remains a highly challenging problem for robots: it demands not only low-level path planning to avoid obstacles and reach the destination, but also high-level scene understanding to interpret and make sense of the surrounding environment.

Most existing navigation methods rely on the construction of prior maps\cite{cadena2017past} or require extensive scene-specific data for task-oriented pre-training\cite{szot2021habitat}. Recently, some works\cite{yin2024sg,zhong2024topv,wu2024voronav} have begun to explore zero-training and zero-shot navigation strategies, relying on textual descriptions of the current task, image inputs, and previously observed historical information, these approaches achieve navigation without dependence on environment or task-specific data, gradually shedding the reliance on scene priors.

Although zero-shot and zero-training navigation methods offer a novel perspective, they still face numerous challenges in practical applications. On the one hand, most current navigation methods are non-end-to-end, where the agent's spatial actions is mapped to a discrete set. These discrete actions result in paths that are neither smooth nor efficient. To align with a target, the agent may require multiple small-angle rotations. On the other hand, the primary bottleneck for current Vision-Language Models (VLMs) in long-range navigation is their inadequate memory mechanisms. Their reliance on discrete observational inputs prevents a cohesive understanding of spatiotemporal continuity. More critically, the prevalent approaches\cite{ramakrishnan2024does,nasiriany2024pivot} of storing history as an unstructured log within a single-step decision paradigm fundamentally compromises their ability to perform effective long-term path planning.


Even though end-to-end methods like VLMnav\cite{nasiriany2024pivot,goetting2024end} utilize historical information, they typically store this information in a flat, unstructured manner, which fundamentally limits their ability to perform long-range navigation. Additionally, VLMs sometimes insufficient understanding of task semantics often leads to poor decision-making, and the lack of reliable check mechanisms for navigation states frequently results in unreliable behaviors such as spinning in place during navigation tasks. 

\begin{figure}
    \centering 
    \includegraphics[width=\textwidth]{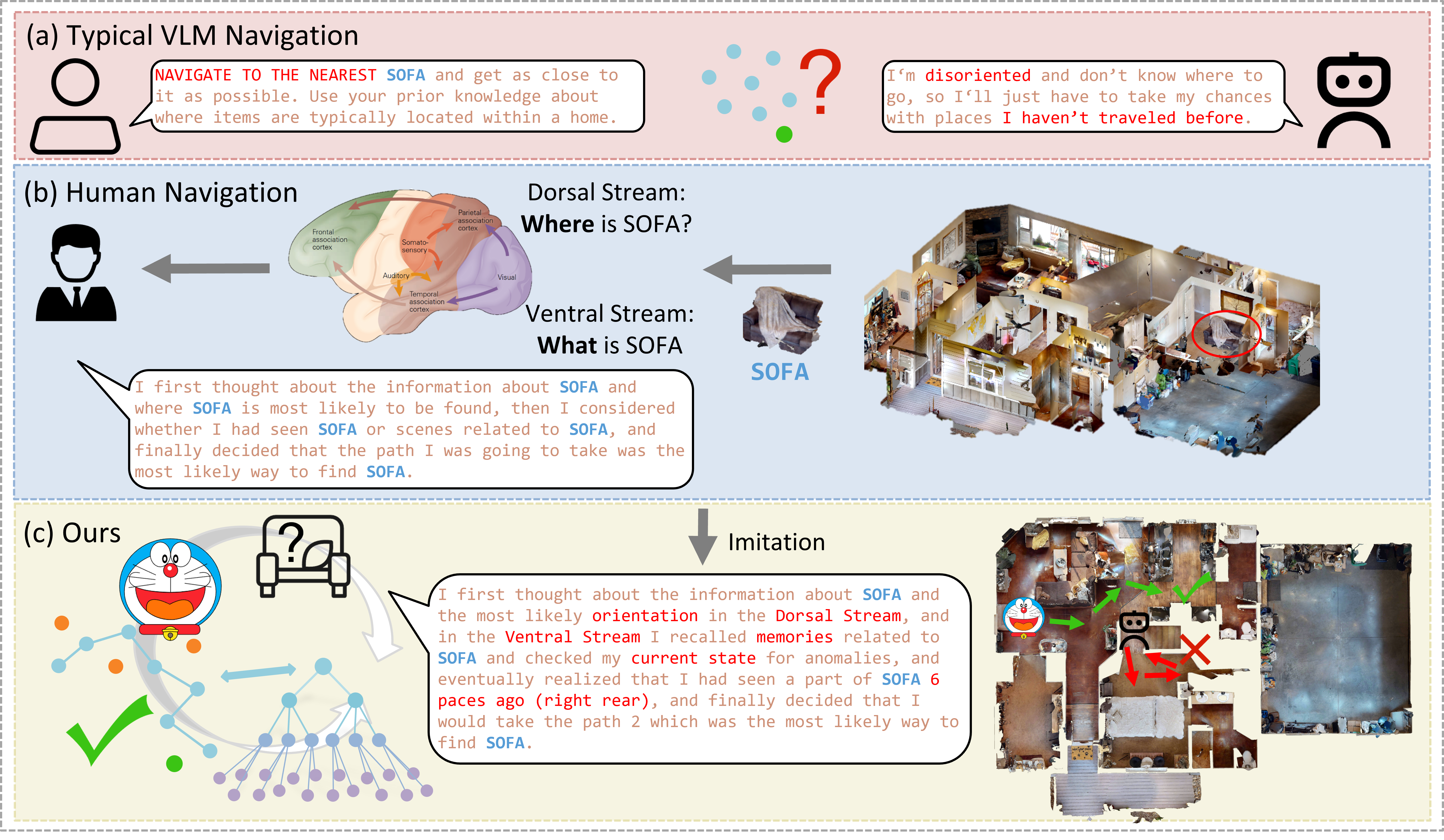} 
    \caption{(a) Illustrates limitation of typical VLM navigation (red arrow). (b) DORAEMON's cognitive inspiration from human navigation. (c) Our DORAEMON method.}
    \label{fig:teaser} 
\end{figure}

Inspired by cognitive science "Decentralized Ontology" principles \cite{bouquet2004contextualizing}, we propose the Decentralized Ontology-aware Reliable Agent with Enhanced Memory Oriented Navigation (DORAEMON), which consists mainly of a Ventral Stream and a Dorsal Stream. The core theoretical premise is that knowledge is inherently distributed and context-dependent, composed of multiple local perspectives, rather than the single, monolithic world model. The Ventral Stream processes object identity (``what'') information, while the Dorsal Stream handles spatial (``where'') processing in the human brain. The Dorsal Stream addresses spatio-temporal discontinuities through a Topology Map and a Hierarchical Semantic-Spatial Fusion, allowing our agent to reason accurately about target-environment relationships. Additionally, the Ventral Stream improves task understanding by utilizing a CoDe-VLM (Compositional Decomposition VLM) and Exec-VLM (Execution VLM) for navigation. Additionally, DORAEMON features a Nav-Ensurance system that enables agent to autonomously detect and respond to abnormal conditions, such as becoming stuck or blocked during navigation. To evaluate navigation performance more comprehensively, we propose a new metric called the Adaptive Online Route Index (AORI). Fig~\ref{fig:teaser} conceptually illustrates limitations of traditional VLN methods and contrasts them with DORAEMON.

In summary, the main contributions of this work are:

\begin{itemize}
    \item We propose DORAEMON, a novel adaptive navigation framework inspired by cognitive principles of decentralized knowledge, consisting of ventral and Dorsal Streams, enabling end-to-end and zero-shot navigation in completely unfamiliar environments without pre-training, offering plug-and-play compatibility with any VLMs.
    \item We propose the Dorsal Stream, which involves designing a Topology Map and a Hierarchical Semantic-Spatial Fusion Network to effectively manage spatio-temporal discontinuities. Additionally, we introduce the Ventral Stream, incorporating a synergistic reasoning component that combines CoDe-VLM for understanding ontological tasks and Exec-VLM for enhanced task comprehension and planning.
    \item We develop Nav-Ensurance, which includes multi-dimensional stuck detection and context-aware escape mechanisms. We propose a new evaluation metric called AORI to quantify the efficiency of the agent’s exploration. Our method demonstrates state-of-the-art performance across various navigation tasks.
\end{itemize}

\section{Related Work}

\subsection{Zero-shot Navigation}
Navigation methods are broadly supervised or zero-shot. Supervised approaches train visual encoders with reinforcement/imitation learning \cite{khandelwal2022simple,maksymets2021thda,ramrakhya2022habitat,chen2022learning} or build semantic maps from training data \cite{zhang2025novel,min2021film,zheng2022jarvis}, struggling with novel scenarios due to data dependency. Zero-shot methods address this using open-vocabulary understanding, increasingly leveraging foundation models like LLMs and VLMs. LLMs provide commonsense reasoning via object-room correlation \cite{yin2024sg,zhou2023esc,wu2024voronav}, semantic mapping \cite{Yu_2023}, and chain-of-thought planning \cite{cai2025cl,yin2024sg,shah2023navigation}, while VLMs align visual observations with textual goals. These foundation model-guided techniques include image-based methods mapping targets to visual embeddings \cite{wen2025zero,gadre2023cows,al2022zero} and map-based approaches using frontier \cite{zhong2024topv,zhang2025novel,chen2023not,kuang2024openfmnav,Yu_2023,shah2023navigation} or waypoint-based maps \cite{wu2024voronav} with LLM/VLM reasoning. VLM-based strategies either use VLMs for recognition with traditional planning and extra perception models \cite{rahmanzadehgervi2024vision,zhang2025apexnav}, or, like PIVOT \cite{nasiriany2024pivot} and VLMnav \cite{goetting2024end}, directly produce actions end-to-end via visual prompting. Despite progress, many zero-shot methods, especially those processing observations independently, face challenges integrating temporal information and handling complex spatial reasoning in unfamiliar environments.

\subsection{Memory Mechanisms in Navigation}
Memory representations in navigation systems have evolved through various architectures, including episodic buffers that maintain observation sequences \cite{goetting2024end,shah2023lm,hsu2022improving}, spatial representations prioritizing geometric information \cite{zhong2024topv,zhang2025apexnav}, graph-based semantic structures capturing object relationships \cite{yin2025unigoal,yin2024sg}, predictive world models attempting to forecast environmental states \cite{cao2024cognav,nie2025wmnav} and the memory capacity acquired through training\cite{zhu2025move}. These systems typically process semantic and spatial information separately, with limited integration between perception and reasoning modules. Most approaches focus on either building representations or enhancing reasoning mechanisms independently. Differently, DORAEMON integrates these aspects through a hierarchical semantic-spatial fusion network with bidirectional information flow between ventral and dorsal processing streams.

\subsection{Cognitive Neuroscience Inspiration in Navigation}

Navigation systems are influenced by cognitive neuroscience, recent models like CogNav\cite{cao2024cognav} and BrainNav\cite{ling2025endowing} incorporate cognitive elements, but they do not fully embody Decentralized Ontology. CogNav utilizes a finite state machine for cognitive states, but may have limitations in knowledge integration. BrainNav mimics biological functions but doesn't deeply engage in decentralized information processing. In contrast, DORAEMON is inspired by Decentralized Ontology\cite{bouquet2004contextualizing}, which suggests that human knowledge is organized through interconnected cognitive systems that enable context-dependent reasoning. It emphasizes the integration and bidirectional exchange of information between Dorsal Stream and Ventral Stream, allowing for the construction of semantic relationships that enhance spatial understanding and support flexible, context-aware navigation.

\section{Methods}

\paragraph{Task Formulation}
We address the Navigation task \cite{DBLP:journals/corr/abs-2006-13171}, where an agent, starting from an initial pose, must locate and navigate to a target within a previously unseen indoor environment. At step $t$, the agent receives observation $I_t$, current pose $P_t$ and a task specification $T$, which can be either a simple object category (e.g., ``sofa'') or an instruction (e.g., ``find the red chair'' or``the plant on the desk'') for tasks like GOAT \cite{khanna2024goat}. Based on these inputs, the agent must decide on an action $a_t$. While many prior works utilize a discrete action space, our end-to-end framework employs a continuous action representation in polar coordinates $(r_t, \theta_t)$, where $r_t$ specifies the forward distance to move, and $\theta_t$ denotes the change in orientation. Crucially, the action space also includes a \texttt{stop} action. The task is considered successful if the agent executes the \texttt{stop} action after meeting successive stop triggers in steps $t$ and $t+1$. The trigger occurs when 1) the agent is within a predefined distance threshold $d_{success}$ of the target object;  2) the target object is visually confirmed within the agent's current observation $I_t$.

\paragraph{Methods Overview}

\begin{figure}
    \centering 
    \includegraphics[width=\textwidth]{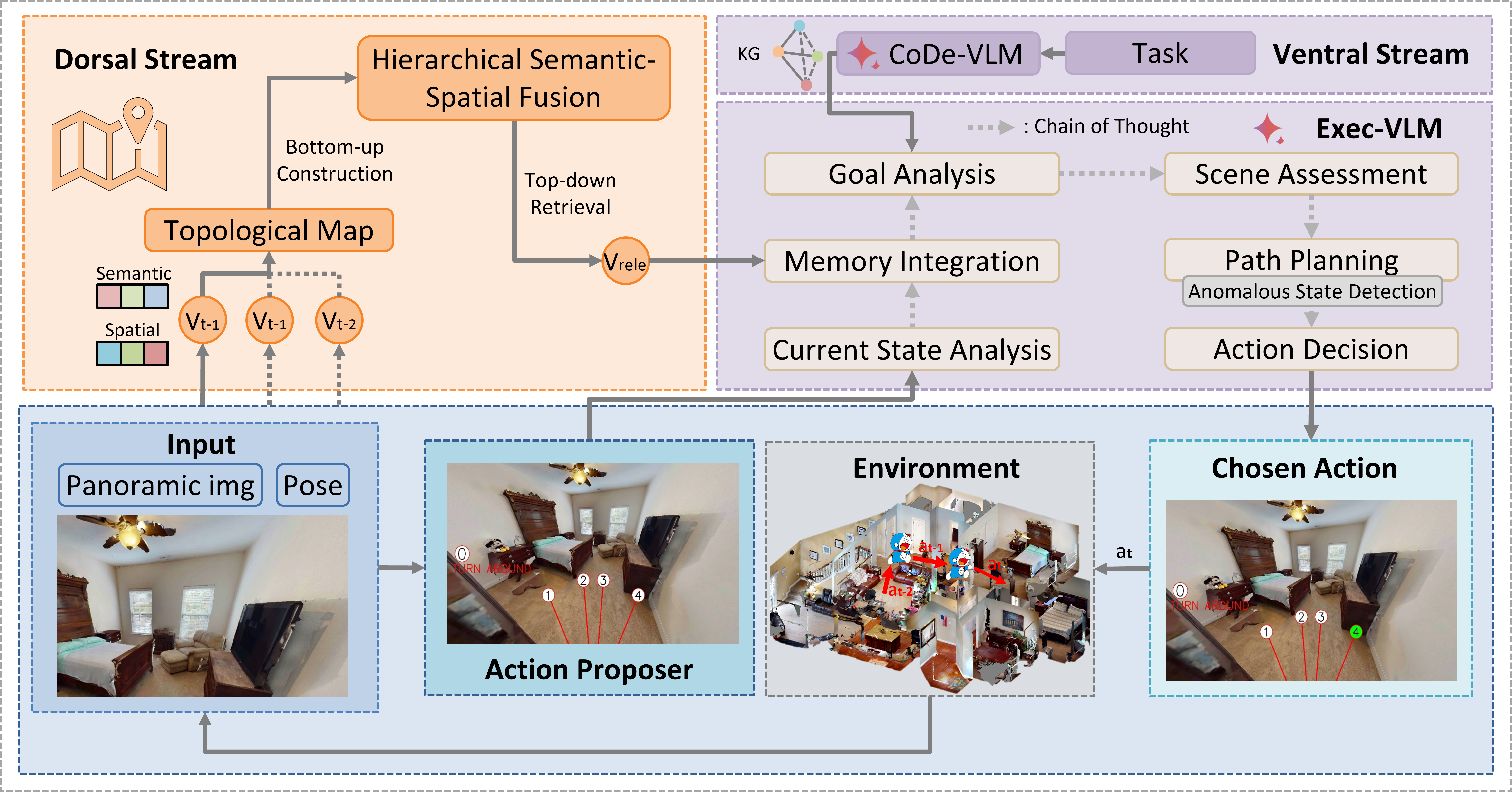} 
    \caption{Architecture of the DORAEMON Navigation Framework.}
    \label{fig:overall} 
\end{figure}

Our DORAEMON framework achieves end-to-end and zero-shot navigation through two decentralized cognitive-inspired streams, as depicted in Figure~\ref{fig:overall}. Given an input with a panoramic image $I_t$ and a pose $P_t$ at step $t$, they are processed by Action Proposer (Appendix~\ref{app:action_proposer}) and Dorsal Stream(Section~\ref{subsec:Dorsal Stream}), respectively. In the Action Proposer, a candidate image $I^t_{\text{anno}}$ is generated with a set of action candidates $A_{\text{final}}^t$. Concurrently, the Dorsal Stream extracts semantic and spatial information from $I_t$ using Hierarchical Semantic-Spatial Fusion and stores it within the Topology Map as node ${v}_t$. The relevant node ${v}_{\text{rele}}$ can be accessed by up-down retrieval. After that, ${v}_{\text{rele}}$ and $I^t_{\text{anno}}$ are input to the Exec-VLM to select the best action based on the given information(Section~\ref {subsubsec:Exec-VLM}). At the same time, the Exec-VLM receives a task-specific knowledge graph (KG) relevant to the task $T$, which is generated by the CoDe-VLM (Section\ref{subsubsec:CoDe-VLM}) in the Ventral Stream (Section~\ref {subsec:Ventral Stream}). The Exec-VLM integrates the information through chain of thought (Appendix~\ref{app:cot_prompt}), identifies abnormal conditions (Section~\ref {subsec:safety_efficiency}), and outputs the final action $a_t$. The agent performs this action $a_t$ in the environment, navigates, and makes the next decision at step $t+1$.

\subsection{Dorsal Stream}\label{subsec:Dorsal Stream}

\begin{figure} 
    \centering 
    \includegraphics[width=\textwidth]{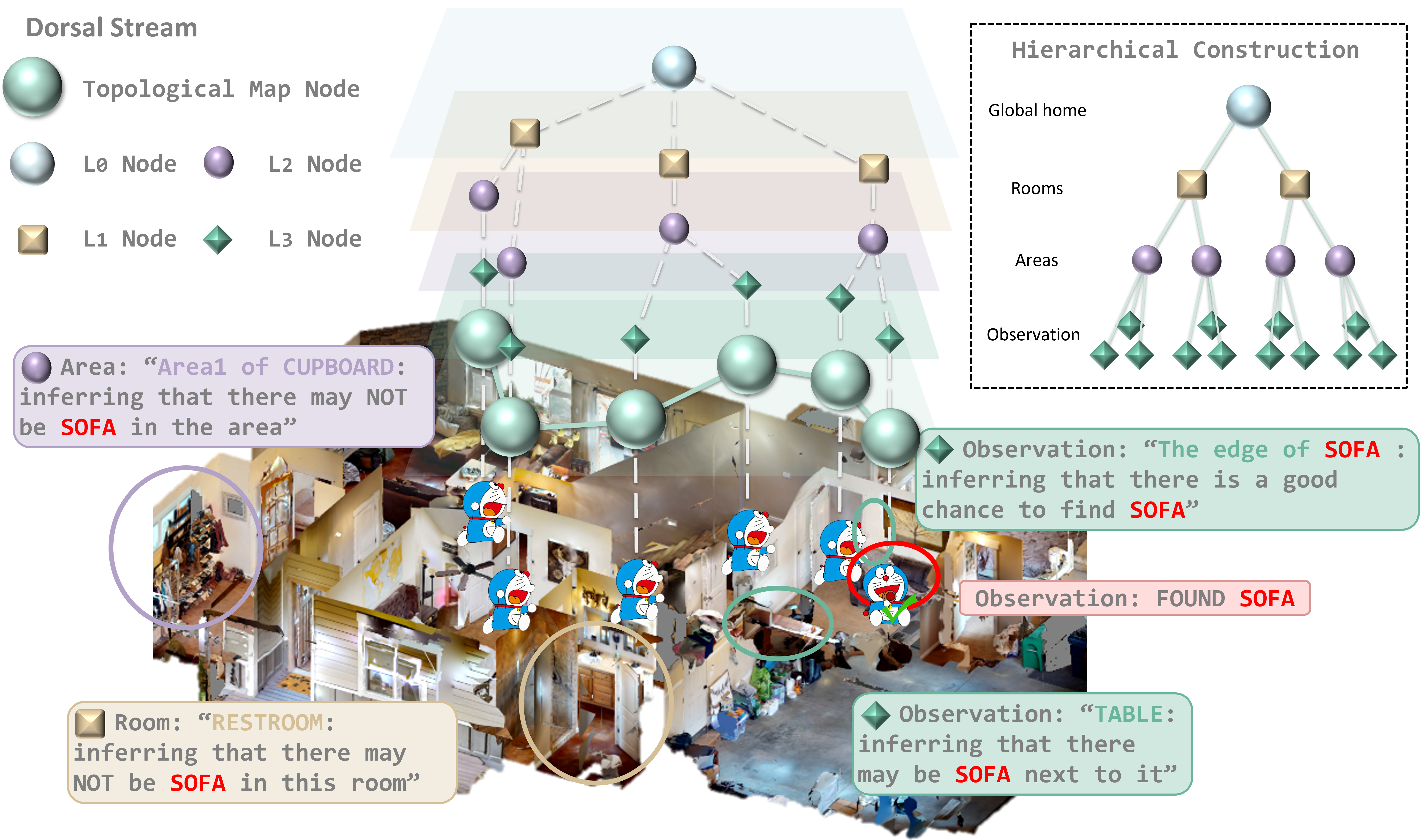} 
    \caption{Architecture of Topological Map and Hierarchical Construction built in Dorsal Stream for spatio-temporal memory. The top view in the middle shows the content of different nodes during navigation, and the upper right part represents the Hierarchical Construction of a node.}
    \label{fig:dorsal_stream} 
\end{figure}

The Dorsal Stream, similar to the ``where/how'' pathway in cognition, is responsible for processing the spatial information to effectively navigate. As illustrated in Figure~\ref{fig:dorsal_stream}, at each step $t$, the agent constructs $v_k$ on the Topology Map (Section~\ref{subsubsec:TopologicalMap}). Subsequently, the Hierarchical Semantic-Spatial Fusion (Section~\ref{subsubsec:SpatialMemoryModule}) organizes the information into a hierarchical structure from the bottom up.

\subsubsection{Topological Map}\label{subsubsec:TopologicalMap}

The topological map $\mathcal{G} = (\mathcal{V}, \mathcal{E})$ is constructed incrementally. Each node $v_t \in \mathcal{V}$ encapsulates the agent's state at timestep $t$, defined as a tuple $(p_t, q_t, I_t, L_t, o_t, s_t)$. $p_t$ and $q_t$ denote the agent's position and orientation, which constitute the pose $P_t$. $I_t$ is the visual observation, $L_t$ is its corresponding language description, $o_t$ is the target likelihood estimation, and $s_t$ represents a semantic embedding of the observation (e.g., CLIP features). A new node $v_{\text{new}}$ is added to $\mathcal{V}$ based on spatio and temporal criteria: a new node is created if either the time elapsed since the last node addition $t_{\text{curr}} - t_{\text{prev}}$ exceeds a temporal threshold $S_{\text{update}}$, or if the agent's Euclidean distance from the previous node $\|{p}_{\text{curr}} - {p}_{\text{prev}}\|_2$ surpasses a spatial threshold $\delta_{\text{sample}}$. Upon its creation, $v_{\text{new}}$ is connected to its predecessor node $v_{\text{prev}}$ via a new edge.

\subsubsection{Hierarchical Semantic-Spatial Fusion}\label{subsubsec:SpatialMemoryModule} 
\textbf{Hierarchical Construction.} 
Building upon the information associated with the Topological Map nodes $v_t \in \mathcal{V}$,  our module organizes information of  $v_t$ into a hierarchical structure. The nodes $h_j$ on the hierarchical structure are defined as:  

\begin{equation}
h_j = \Big( id_j,\ l_j,\ \mathcal{P}_j,\ \mathcal{C}_j \Big), 
\label{eq:memory_hierarchy}
\end{equation}

where $id_j$, $l_j \in \{L_0, L_1, L_2, L_3\}$, $\mathcal{P}_j$, $\mathcal{C}_j$ correspond to unique string identifier, hierarchy level tag, parent node references, and child node references.

The memory hierarchy organizes nodes $h_j$ into four semantic levels through structural and functional relationships (Appendix~\ref{app:Hierarchical Construction}):  $L_3$ (Observation, directly linked to topological map nodes $v_t$), $L_2$ (Area), $L_1$ (Room), $L_0$ (Environment).  
The memory hierarchy is constructed bottom-up ($L_3 \to L_2 \to L_1 \to L_0$) after an initial exploration phase or periodically. While the overall process involves sequential clustering or integration steps for each level transition, the specific logic and parameters differ between levels.

\textbf{Hierarchical Memory Retrieval.} 
To efficiently find relevant information within the constructed hierarchy (e.g., observations related to sofa), the system employs a top-down search. This search is guided by a scoring function $S(h_i)$ evaluated at nodes $h_i$ during traversal the constructed hierarchy:

\begin{gather}
S(h_i) = \alpha_{\text{sem}} S_{\text{sem}}(h_i, T) + \alpha_{\text{spa}} S_{\text{spa}}(h_i) + \alpha_{\text{key}} S_{\text{key}}(h_i, T) + \alpha_{\text{time}} S_{\text{time}}(h_i), \label{eq:scoring_function_revised} 
\end{gather}

where $S_{\text{sem}}$ computes embedding similarity between node $h_i$ and task $T$, $S_{\text{spa}}$ measures proximity to the current position using an exponential decay function, $S_{\text{key}}$ evaluates keyword overlap, and $S_{\text{time}}$ prioritizes recent observations. For instance, semantic similarity is calculated via normalized cosine similarity, while spatial and temporal scores both rely on exponential decay models to reflect their diminishing influence over distance and time. (see more details in Appendix \ref{app:scoring_details}).




\subsection{Ventral Stream}\label{subsec:Ventral Stream}
The Ventral Stream, analogous to the ``what'' pathway in human cognition, integrates two key components: CoDe-VLM (Compositional Decomposition VLM, Section~\ref{subsubsec:CoDe-VLM}) and Exec-VLM (Exection VLM, Section~\ref{subsubsec:Exec-VLM}). Unlike prior models that encode task information into a single, entangled vector, our architecture explicitly disentangles task comprehension from execution. This decentralized design mirrors the Ventral Stream's approach to compositional understanding, first compiling knowledge and then acting upon it.

\subsubsection{CoDe-VLM: Compiling Tasks into Knowledge Graphs}
\label{subsubsec:CoDe-VLM}
To build a deep and structured understanding of the task, CoDe-VLM acts as a semantic compiler. It leverages the vast world knowledge embedded within a VLM to on-the-fly compile an unstructured instruction $T$ into a dynamic, task-specific knowledge graph (KG). 



CoDe-VLM generates a graph structure encapsulating nodes and relational edges. This task KG, formed from extracted semantic attributes like general description, appearance, and location, constitutes our explicit and compositional representation of the task. This representation not only enables the agent to robustly verify objects encountered during navigation but also provides crucial priors for planning by interfacing with the spatial reasoning components of the Dorsal Stream.

\subsubsection{Exec-VLM: Executing Actions via Graph-based Reasoning}
\label{subsubsec:Exec-VLM}
The Exec-VLM serves as the agent's executive core, responsible for determining the optimal action by combining visual observations, spatial awareness from Dorsal Stream, and the structured task semantics provided by CoDe-VLM. Crucially, instead of making decisions in a high-dimensional, entangled feature space, Exec-VLM performs explicit reasoning on the task knowledge graph.

We steer this reasoning process using Chain-of-Thought (CoT). The CoT guides Exec-VLM to break down the complex navigation task into interpretable sub-steps: current state analysis, memory integration, goal analysis, scene assessment, path planning, and action decision. During the ``goal analysis'' step, for instance, the model directly queries the nodes and edges of the KG to confirm the target's identity and properties, rather than relying on a fragile memory of the initial instruction.


\subsection{Nav-Ensurance}\label{subsec:safety_efficiency} 

To enhance the evaluation of safety and efficiency in navigation, we present a new metric Area Overlap Redundancy Index (AORI) (Section~\ref{subsubsec:aori}). Additionally, we develop Nav-Ensurance, including Multimodal Stuck Detection (Section~\ref {subsubsec:stuck_detection}), context-aware escape strategies (Section~\ref {subsubsec:escape}), and adaptive precision navigation (Section~\ref {subsubsec:precision}) to ensure navigation systems reliably and effectively.

\subsubsection{Area Overlap Redundancy Index (AORI)}\label{subsubsec:aori}
We introduce the Area Overlap Redundancy Index (AORI) to quantify the efficiency of the agent's navigation strategy by measuring overlap in area coverage. A high AORI indicates excessive path overlap and inefficient exploration,  specifically addressing the limitations of conventional coverage metrics that neglect temporal-spatial redundancy. AORI is formally defined as:

\begin{equation}
    \text{AORI} = 1.0 - (w_c \cdot (1.0 - r_{\text{overlap}})^2 + w_d \cdot (1.0 - d_{\text{norm}})), \label{eq:aori}
\end{equation}

Where $r_{\text{overlap}}$ represents the ratio of revisited areas, $d_{\text{norm}}$ is the normalized density, and $w_c = 0.8, w_d = 0.2$ are weighting coefficients. Further details are provided in Appendix~\ref{app:aori}.

\subsubsection{Multimodal Stuck Detection}
\label{subsubsec:stuck_detection}

To detect if it is stuck, the agent analyzes its trajectory over a sliding window of $T$ steps by computing two key metrics: the progress efficiency $\eta$ and the rotational-to-translational ratio $\rho$.
\begin{equation}
\eta = \frac{\|{p}_T - {p}_0\|_2}{\sum_{t=1}^{T} \|{p}_t - {p}_{t-1}\|_2}, \quad \rho = \frac{\sum_{t=1}^{T} |\theta_t - \theta_{t-1}|}{\sum_{t=1}^{T} \|{p}_t - {p}_{t-1}\|_2}.
\label{eq:stuck_metrics}
\end{equation}
A stuck state is confirmed if a weighted score $S = w_\eta \cdot \mathbb{I}[\eta < \tau_\eta] + w_\rho \cdot \mathbb{I}[\rho > \tau_\rho]$ remains above a threshold $S_{\text{th}}$ for $k$ consecutive windows. These metrics effectively detect situations where the agent makes little forward progress (low $\eta$) or is spinning in place (high $\rho$).

\subsubsection{Context-aware Escape Strategies}\label{subsubsec:escape}
When a stuck state is detected, the system selects an appropriate escape strategy based on the perceived information from Dorsal Stream(Section~\ref{subsec:Dorsal Stream}). For instance, in corner traps (perceived dead ends), a large turn is executed. In narrow passages, a small backward step followed by a randomized direction change is employed. If the environmental context is ambiguous, the agent will analyze recent successful movement directions and attempt to move perpendicularly, significantly improving escape capabilities from complex trap situations.

\subsubsection{Adaptive Precision Navigation}\label{subsubsec:precision}
As the agent nears the target object, it will activate a precision navigation mode. In this mode, the distance component $d$ of all proposed actions $(d, \theta)$ is scaled down by a factor $\gamma_{\text{step}}$ to enable fine-grained positioning adjustments:
\begin{equation}
    a_{\text{precise}} = (d \cdot \gamma_{\text{step}}, \theta) \quad \text{for action } (d, \theta) \in A_{\text{actions}}. \label{eq:precision_full}
\end{equation}
Additionally, when activating the precision navigation mode, the system can utilize visual analysis (using VLM) to create more detailed action options, thereby maximizing final positioning accuracy.

\section{Experiments}\label{sec:experiments}

\paragraph{Datasets}
We evaluate our proposed DORAEMON within the Habitat simulator \cite{habitat19iccv} on four large-scale datasets: HM3Dv1\cite{ramakrishnan2021hm3d}(Object Navigation), HM3Dv2\cite{yadav2023habitat}(Object Navigation), and MP3D \cite{chang2017matterport3d}(Object Navigation), GOAT\cite{khanna2024goat} (Multi-modal lifelong
navigation, using HM3Dv2).

\paragraph{Implement Details and Evaluation Metrics}
The action space includes \texttt{stop}, \texttt{move\_forward} where the distance parameter is sampled from the continuous range $[0.5\text{m}, 1.7\text{m}]$, and \texttt{rotate}. 
We adopt standard metrics to evaluate navigation performance: Success Rate (SR), the percentage of episodes where the agent successfully stops near a target object; Success weighted by Path Length (SPL), defined as $\frac{1}{N}\sum_{i=1}^{N}S_i\frac{l_i}{\max(p_i, l_i)}$, rewarding both success and efficiency; and our proposed Area Overlap Redundancy Index (AORI) (Equation~\eqref{eq:aori}), which quantifies navigation by penalizing redundant exploration (lower is better). More information is set in the Appendix~\ref{app:setup_structured}.


\paragraph{Baselines}
We compare DORAEMON against several state-of-the-art navigation methods on the HM3Dv2, HM3Dv1, and MP3D. Our main comparison focuses on end-to-end approaches. Beyond these direct end-to-end counterparts, we also consider a broader set of recent methods for non-end-to-end object navigation methods. More baseline details are set in the Appendix~\ref{app:basline}.


\subsection{Methods Comparision}\label{subsec:methods comparision}

\paragraph{End-to-end Methods:}
We evaluate our approach on the HM3Dv2 (ObjectNav,val, Table~\ref{tab:combined_comparison} (a)) and HM3Dv1(GOAT, val, Table~\ref{tab:combined_comparison} (b)) with other end-to-end baselines. DORAEMON achieves state-of-the-art performance on both datasets, outperforming other methods by a significant margin. 

\captionsetup[table]{font=small}  
\begin{table}[htbp]
  \caption{Comparison of end-to-end navigation methods on different benchmarks.}
  \label{tab:combined_comparison}
  \centering
  \begin{minipage}{0.48\textwidth}
    \centering
    \scriptsize
    \caption*{(a) ObjectNav benchmark}
    \resizebox{\textwidth}{!}{
    \begin{tabular}{lccc}
      \toprule
      Method & SR (\%) $\uparrow$ & SPL (\%) $\uparrow$ & AORI (\%) $\downarrow$ \\ 
      \midrule
      Prompt-only & 29.8 & 0.107 & - \\
      PIVOT\cite{nasiriany2024pivot} & 24.6 & 10.6 & 63.3 \\
      VLMNav\cite{goetting2024end} & 51.6 & 18.3 & 61.5 \\
      \textbf{DORAEMON (Ours)} & \textbf{62.0} & \textbf{23.0} & \textbf{50.1} \\
      \midrule
      \textbf{Improvement} & \textbf{20.2} & \textbf{10.0} & \textbf{18.5} \\
      \bottomrule
    \end{tabular}
    }
  \end{minipage}%
  \hfill
  \begin{minipage}{0.48\textwidth}
    \centering
    \scriptsize
    \caption*{(b) GOAT benchmark}
    \resizebox{\textwidth}{!}{
    \begin{tabular}{lccc}
      \toprule
      Method & SR (\%) $\uparrow$ & SPL (\%) $\uparrow$ & AORI (\%) $\downarrow$ \\ 
      \midrule
      Prompt-only & 11.3 & 3.7 & - \\
      PIVOT\cite{nasiriany2024pivot} & 8.3 & 3.8 & 64.9 \\
      VLMNav\cite{goetting2024end} & 22.1 & 9.3 & 63.6 \\
      \textbf{DORAEMON (Ours)} & \textbf{24.3} & \textbf{10.3} & \textbf{56.9} \\
      \midrule
      \textbf{Improvement} & \textbf{10.0} & \textbf{10.8} & \textbf{10.5} \\
      \bottomrule
    \end{tabular}
    }
  \end{minipage}
\end{table}

\paragraph{Non-end-to-end methods:}
Most methods are non-end-to-end, their reliance on fine-grained discrete actions is a significant departure from natural human behavior, underscoring the superiority of an end-to-end approach. To ensure a fair comparison with these methods that utilize a discrete action set $\mathcal{A}$: \texttt{move forward} 0.25m, \texttt{turn left/turn right} 30$^\circ$, \texttt{look up/lookdown} 30$^\circ$, \texttt{stop}, and a common 500 steps episode limit, we conduct an additional set of experiments. In these, we normalize our agent's interactions by approximating an equivalent number of standard discrete steps for each of DORAEMON's actions. During our experiments, one DORAEMON step $t$ was equivalent to about 9-10 non-end-to-end step $t_n$.

\definecolor{MyGray}{gray}{0.9}
\definecolor{MyHighlight}{gray}{0.8}

\begin{table*}[htbp]
  \caption{Comprehensive comparison with state-of-the-art methods on ObjectNav benchmarks. TF refers to training-free, ZS refers to zero-shot, and E2E refers to end-to-end. }
  \label{tab:comprehensive_comparison_styled}
  \centering
  \scriptsize
  \setlength{\tabcolsep}{3pt}
  \rowcolors{3}{white}{MyGray}

  \begin{tabularx}{\linewidth}{>{\raggedright\arraybackslash}Xccccccccc}
    \toprule
    Method & 
    \multicolumn{1}{c}{\makecell{ZS}} & 
    \multicolumn{1}{c}{\makecell{TF}} & 
    \multicolumn{1}{c}{\makecell{E2E}} & 
    \multicolumn{2}{c}{HM3Dv1} & 
    \multicolumn{2}{c}{HM3Dv2} & 
    \multicolumn{2}{c}{MP3D} \\
    \cmidrule(lr){5-6} \cmidrule(lr){7-8} \cmidrule(lr){9-10}
    & & & & 
    \makecell{SR(\%) $\uparrow$} & \makecell{SPL(\%) $\uparrow$} & 
    \makecell{SR(\%) $\uparrow$} & \makecell{SPL(\%) $\uparrow$} & 
    \makecell{SR(\%) $\uparrow$} & \makecell{SPL(\%) $\uparrow$} \\
    \midrule
    ProcTHOR \cite{deitke2022} & 
    $\times$ & $\times$ & $\times$ & 54.4 & \textbf{31.8} & - & - & - & - \\
    SemEXP \cite{chaplot2020object}& 
    $\checkmark$ & $\times$ & $\times$ & - & - & - & - & 36.0 & 14.4 \\ 
    Habitat-Web\cite{ramrakhya2022habitat}  & 
    $\checkmark$ & $\times$ & $\times$ & 41.5 & 16.0 & - & - & 31.6 & 8.5 \\ 
    PONI \cite{ramakrishnan2022poni} & 
    $\checkmark$ & $\times$ & $\times$ & - & - & - & - & 31.8 & 12.1 \\     
    ProcTHOR-ZS \cite{deitke2022} & 
    $\checkmark$ & $\times$ & $\times$ & 13.2 & 7.7 & - & - & - & - \\
    ZSON \cite{majumdar2022zson} & 
    $\checkmark$ & $\times$ & $\times$ & 25.5 & 12.6 & - & - & 15.3 & 4.8 \\
    PSL \cite{sun2024prioritized} & 
    $\checkmark$ & $\times$ & $\times$ & 42.4 & 19.2 & - & - & - & - \\
    Pixel-Nav \cite{cai2024bridging} & 
    $\checkmark$ & $\times$ & $\times$ & 37.9 & 20.5 & - & - & - & - \\
    SGM \cite{zhang2024imagine} & 
    $\checkmark$ & $\times$ & $\times$ & \textbf{60.2} & 30.8 & - & - & 37.7 & 14.7 \\
    ImagineNav \cite{zhao2024imaginenav} & 
    $\checkmark$ & $\times$ & $\times$ & 53.0 & 23.8 & - & - & - & - \\
    CoW \cite{gadre2023cows} & 
    $\checkmark$ & $\checkmark$ & $\times$ & - & - & - & - & 7.4 & 3.7 \\
    ESC \cite{zhou2023esc} & 
    $\checkmark$ & $\checkmark$ & $\times$ & 39.2 & 22.3 & - & - & 28.7 & 14.2 \\
    L3MVN \cite{Yu_2023} & 
    $\checkmark$ & $\checkmark$ & $\times$ & 50.4 & 23.1 & 36.3 & 15.7 & 34.9 & 14.5 \\
    VLFM \cite{yokoyama2024vlfm} & 
    $\checkmark$ & $\checkmark$ & $\times$ & 52.5 & 30.4 & 63.6 & \textbf{32.5} & 36.4 & \textbf{17.5} \\    
    VoroNav \cite{wu2024voronav} & 
    $\checkmark$ & $\checkmark$ & $\times$ & 42.0 & 26.0 & - & - & - & - \\
    TopV-Nav \cite{zhong2024topv} & 
    $\checkmark$ & $\checkmark$ & $\times$ & 52.0 & 28.6 & - & - & 35.2 & 16.4 \\
    InstructNav \cite{long2024instructnav} & 
    $\checkmark$ & $\checkmark$ & $\times$ & - & - & 58.0 & 20.9 & - & - \\  
    SG-Nav \cite{yin2024sg} & 
    $\checkmark$ & $\checkmark$ & $\times$ & 54.0 & 24.9 & 49.6 & 25.5 & 40.2 & 16.0 \\  
    \midrule
    \rowcolor{MyHighlight}
    \textbf{DORAEMON (Ours)} & 
    $\checkmark$ & $\checkmark$ & $\checkmark$ & 
    55.6 & 21.4 & \textbf{66.5} & 20.6 & \textbf{41.1} & 15.8 \\   
    \bottomrule
  \end{tabularx}
\end{table*}

Compared to the non-end-to-end approach in the Table~\ref{tab:comprehensive_comparison_styled}, DORAEMON achieves state-of-the-art performance on SR, despite normalizing our action to set $\mathcal{A}$. Each action performed by ours corresponds to several actions in this set. In fact, we only run about 60 end-to-end steps, which further demonstrates the excellence of our DORAEMON.(details are provided in the Appendix~\ref{app:action trans})

\paragraph{Ablation Studies.}
We perform comprehensive ablation studies to validate our design choices, with results summarized in Table~\ref{tab:ablation_ultimate_styled}. (a) Core Components: Table~\ref{tab:ablation_ultimate_styled}(a) shows that each component is crucial. Removing both the Dorsal and Ventral streams severely degrades performance, confirming their synergistic effect. Disabling the Nav-Ensurance mechanism also notably worsens the AORI, highlighting its effectiveness in error prevention. (b) Choice of VLM: The VLM ablation in Table~\ref{tab:ablation_ultimate_styled}(b) indicates that while Gemini-1.5-Pro is optimal, our framework remains highly effective with smaller models. This demonstrates our architecture's inherent strength and its plug-and-play nature, suggesting future compatibility with evolving VLMs. (c) Hyperparameter Sensitivity: The analysis in Table~\ref{tab:ablation_ultimate_styled}(c) reveals a trade-off between metrics. For example, setting TopK=12 yields the highest SR but at the cost of SPL. Our default hyperparameters are carefully chosen to achieve a robust and balanced performance across all metrics, rather than over-optimizing for a single one.

\begin{table}[htbp]
  \centering
  \caption{A comprehensive ablation study of DORAEMON across different datasets, including variations in modules, VLMs, and hyperparameters. All experiments were evaluated over 100 episodes. Our DORAEMON uses the default hyperparameters: TopK=8, memory update interval=3, area grid size=2.}
  \label{tab:ablation_ultimate_styled}
  \small
  \resizebox{\textwidth}{!}{%
  \begin{tabular}{l ccc ccc ccc}
    \toprule
    \textbf{Method / Configuration} & \multicolumn{3}{c}{\textbf{HM3Dv2}} & \multicolumn{3}{c}{\textbf{HM3Dv1}} & \multicolumn{3}{c}{\textbf{MP3D}} \\
    \cmidrule(lr){2-4} \cmidrule(lr){5-7} \cmidrule(lr){8-10}
     & SR (\%) $\uparrow$ & SPL (\%) $\uparrow$ & AORI (\%) $\downarrow$ & SR (\%) $\uparrow$ & SPL (\%) $\uparrow$ & AORI (\%) $\downarrow$ & SR (\%) $\uparrow$ & SPL (\%) $\uparrow$ & AORI (\%) $\downarrow$ \\
    \midrule
    \multicolumn{10}{c}{\cellcolor{gray!20}\textit{(a) Ablation of different modules}} \\
    \midrule
    w/o Dorsal \& Ventral Stream & 51.6 & 18.3 & 61.5 & 48.4 & 18.9 & 53.7 & 38.8 & 13.9 & 64.3 \\
    w/o Dorsal \& CoDe-VLM      & 54.0 & 19.8 & 59.1 & 51.2 & 19.4 & 52.5 & 40.2 & 14.2 & 63.8 \\
    w/o Dorsal Stream           & 59.0 & 22.7 & 56.3 & 53.8 & 20.5 & 51.1 & 40.9 & 14.6 & 65.1 \\
    w/o Nav-Ensurance           & 60.0 & 22.5 & 54.9 & 53.1 & 20.7 & \textbf{50.9} & 42.2 & 15.3 & 60.4 \\
    \midrule
    \multicolumn{10}{c}{\cellcolor{gray!20}\textit{(b) Ablation of different VLMs (on HM3Dv2)}} \\
    \midrule
    Qwen-7B                     & 49.5 & 20.6 & 68.7 & - & - & - & - & - & - \\
    Gemini-1.5-Flash            & 58.0 & 20.1 & 54.8 & - & - & - & - & - & - \\
    Gemini-2-Flash              & 59.0 & 21.5 & 57.9 & - & - & - & - & - & - \\
    \midrule
    \multicolumn{10}{c}{\cellcolor{gray!20}\textit{(c) Ablation of Hyperparameters (on HM3Dv2)}} \\
    \midrule
    \quad w/ TopK = 12                 & \textbf{65.0} & 22.57 & 42.89 & - & - & - & - & - & - \\
    \quad w/ TopK = 4                  & 59.0 & 22.78 & 42.03 & - & - & - & - & - & - \\
    \quad w/ memory update interval = 1 & 53.0 & 19.94 & 45.54 & - & - & - & - & - & - \\
    \quad w/ memory update interval = 5 & 62.0 & 23.61 & 44.22 & - & - & - & - & - & - \\
    \quad w/ area grid size = 1        & 61.0 & 21.97 & \textbf{39.66} & - & - & - & - & - & - \\
    \quad w/ area grid size = 3        & 61.0 & 22.95 & 43.57 & - & - & - & - & - & - \\
    \midrule
    \textbf{DORAEMON (Ours, default)} & 61.0 & \textbf{23.7} & 48.8 & \textbf{55.6} & \textbf{21.4} & \textbf{49.1} & 41.1 & \textbf{15.8} & \textbf{59.3} \\
    \bottomrule
  \end{tabular}
  }
\end{table}

\subsection{Navigation in Real World}
\begin{figure}[h]
  \centering
  \includegraphics[width=1.0\textwidth]{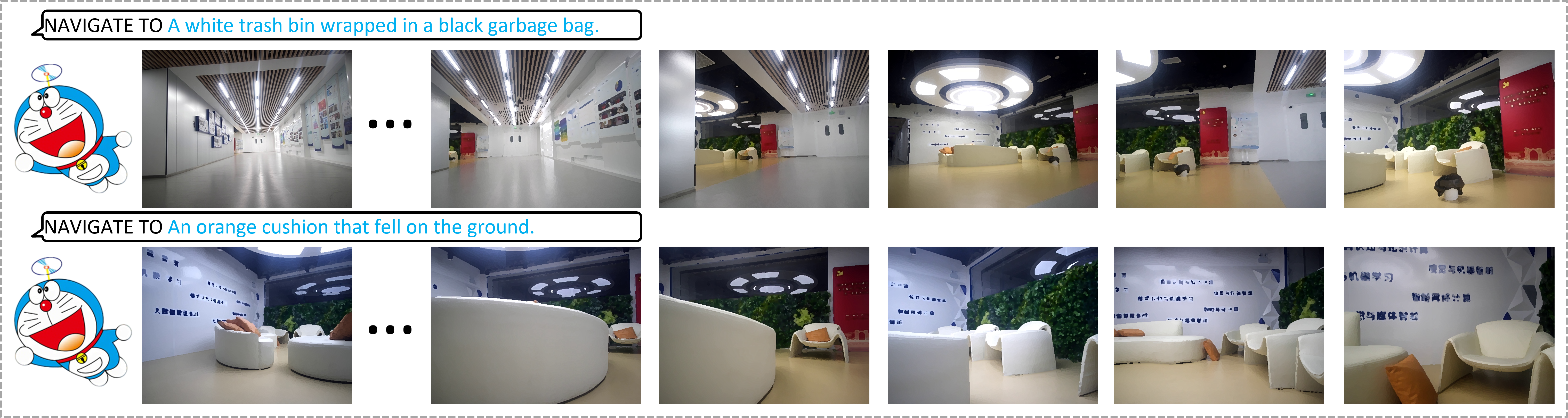}
  \caption{DORAEMON's Performance in sim2real}
  \label{fig:sim2real}
\end{figure}

To validate the Sim-to-Real generalization of our model, we deployed our DORAEMON in a novel office environment.  Despite the significant domain gap, the agent successfully completed navigation tasks. Figure~\ref{fig:sim2real} shows a representative trial. More demos are available on our project homepage.
\section{Conclusion}
In this paper, we present DORAEMON , a novel cognitive-inspired framework consisting of Ventral and Dorsal Streams that mimics human navigation capabilities. The Dorsal Stream implements the Hierarchical Semantic-Spatial Fusion and Topology Map to handle spatiotemporal discontinuities, while the Ventral Stream combines CoDe-VLM and Exec-VLM to improve decision-making. Our approach also develops Nav-Ensurance to ensure navigation safety and efficiency. Extensive experimental results demonstrate the superior performance of DORAEMON.

\newpage
\medskip  
\small 
\bibliographystyle{plain}
\bibliography{references}

\newpage

\appendix
\section{Action Proposer}\label{app:action_proposer}
\begin{figure}[h]
  \centering
  \includegraphics[width=1.0\textwidth]{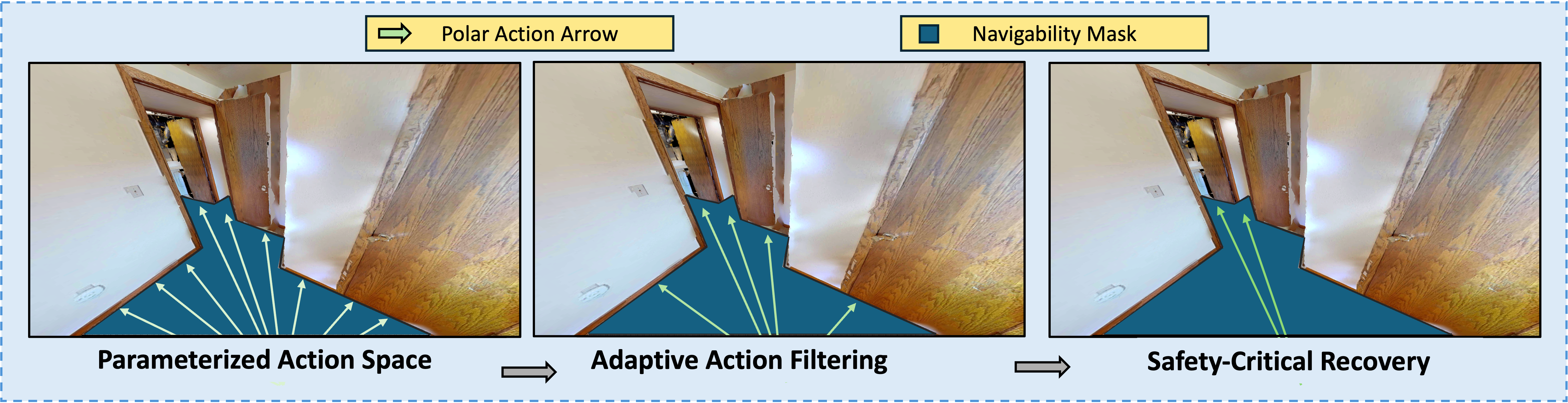}
  \caption{Action proposal: (a) Collision-free action generation within $\pm\theta_{\text{max}}$ FOV, (b) Exploration-aware filtering with $\Delta\theta$ angular resolution, (c) Safety-constrained and action projection.}
  \label{fig:proposer}
\end{figure}

DORAEMON employs an Action Proposer\cite{goetting2024end} to generate a refined set of candidate actions, which the Exec-VLM then evaluates for the final action decision.
As shown in Figure~\ref {fig:proposer}, first parameterized action candidates $A_{\text{init}}^t$ are generated by the parameterized action space (Equation~\eqref{eq:action space}). Second, adaptive filtering (Equation~\eqref{eq:action filter}) refines $A_{\text{cand}}^t$ using exploration state $\mathcal{V}_t$ and historical patterns $\mathcal{H}_t$. Safety-critical recovery (Equation\eqref{eq:safety recovery}) enforces a rotation cooldown $\gamma$ through viability evaluation $\mathcal{F}(\cdot)$. Finally, the projection module visually encodes $A_{\text{final}}^t$ into $I^t_{\text{anno}}$ with numeric tagging (0 for rotation) to interface with VLM's semantic space.

\paragraph{Parameterized Action Space} Define the action space through symbolic parameters:
\begin{equation}
A_{\text{init}}^t = \left\{ \left(\theta_i, \min\left(\eta r_i, r_{\text{max}} \right) \right) \,\bigg|\, \theta_i = k \Delta\theta,\, k \in \mathcal{K} \right\}. 
\label{eq:action space}
\end{equation}
where $\mathcal{K} = \left[-\lfloor\theta_{\text{max}}/\Delta\theta\rfloor, \lfloor\theta_{\text{max}}/\Delta\theta\rfloor\right]$ ensures full FOV coverage. The safety margin $\eta$ and collision check are derived from depth-based navigability analysis.

\paragraph{Adaptive Action Filtering} Refinement combines exploration state $\mathcal{V}_t$ and historical search patterns $\mathcal{H}_t$:
\begin{equation}
A_{\text{cand}}^t = \left\{ (\theta_i,r_i) \in A_{\text{init}}^t \,\bigg|\, \begin{aligned} 
&\alpha(\mathcal{H}_t) \cdot s(\mathcal{V}_t) > \tau, &\min_{\theta_j \in A_{\text{cand}}}|\theta_i-\theta_j| \geq \theta_{\delta}. 
\end{aligned} \right\}
\label{eq:action filter}
\end{equation}
where $\alpha(\cdot)$ models temporal search impact and $s(\cdot)$ quantifies spatial exploration potential.

\paragraph{Safety-Critical Recovery} The next action set enforces,  where $\mathcal{F}(\cdot)$ evaluates action viability and $\gamma$ controls rotation cool down: 
\begin{equation}
A_{\text{final}}^t = \begin{cases} 
\{(\pi,0)\} & \text{if }, \mathcal{F}(A_{\text{cand}}^t) \land (t-t_{\text{rot}} > \gamma) \\
A_{\text{cand}}^t & \text{otherwise}. 
\end{cases}
\label{eq:safety recovery}
\end{equation}

\paragraph{Action Projection} The following phase focuses on visually anchoring these operational elements within the comprehensible semantic realm of the VLM. The projection component annotated visual depiction $I^t_{\text{anno}}$ from $A_{\text{final}}^t$ and $I_t$. We use numeric encoding, assigning a distinct code to each actionable option that is displayed on the visual interface. It is worth noting that rotation is assigned the code 0.

\section{Steps Conversion}\label{app:action trans}

To establish temporal equivalence between DORAEMON's continuous actions and Habitat's discrete steps, we implement the conversion protocol formalized in Algorithm~\ref{alg:step_conversion}. Given a polar action $\mathbf{a} = (r, \theta) \in \mathbb{R}^+ \times (-\pi, \pi]$ with radial displacement $r$ meters and angular rotation $\theta$ radians:

This formulation enables direct comparison with baseline methods by normalizing both:
\begin{equation}
T_{\text{episode}} = \sum_{t=1}^{500} t_n \leq 500
\end{equation}
where $t_n$ denotes converted steps for action at time step $t$. During our experiments, one DORAEMON step $t$ was equivalent to about 9-10 $t_n$

\begin{algorithm}[t]
\caption{Discrete Step Conversion}\label{alg:step_conversion}
    \begin{algorithmic}[1]
    \Require{Polar action $(r, \theta)$, displacement unit $\Delta_r = 0.25$m, angular unit $\Delta_\theta = 30^\circ$}
    \Statex
    \If{action is \texttt{stop}}
        \State \Return $1$ \Comment{Explicit stop handling}
    \Else
        \State $s_r \gets \lceil r / \Delta_r \rceil$ \Comment{Radial step calculation}
        \State $\theta_{\text{deg}} \gets 180|\theta|/\pi$ \Comment{Radian-degree conversion}
        \State $s_\theta \gets \lceil \theta_{\text{deg}} / \Delta_\theta \rceil$ \Comment{Angular step calculation}
        \State $N \gets \max(s_r + s_\theta, 1)$ \Comment{Step composition}
        \State \Return $N$
    \EndIf
    \end{algorithmic}
\end{algorithm}

We also presented examples of numerical conversions for steps in the experiment.

\captionsetup[table]{font=small}  
\begin{table}[htbp]
  \caption{Steps Conversion} 
  \label{tab:steps_conversion_updated} 
  \centering
  \scriptsize 

  \begin{minipage}{0.6\textwidth} 
    \centering
    \caption*{(a) Steps Conversion for an end-to-end Step} 
    \resizebox{\textwidth}{!}{ 
    \begin{tabular}{lcc}
      \toprule
      End-to-end Action & Non-end-to-end Steps & Non-end-to-end Action \\ 
      \midrule
      (1.27m, $53^{\circ}$) & 9 & (1.5m, $75^{\circ}$)/0.25m $\times$ 6 + $25^{\circ}$ $\times$ 3 \\
      (1.7m, $60^{\circ}$)  & 10 & (1.75m, $75^{\circ}$)/0.25m $\times$ 7 + $25^{\circ}$ $\times$ 3 \\
      (1.1m, $93^{\circ}$)  & 9 & (1.25m, $100^{\circ}$)/0.25m $\times$ 5 + $25^{\circ}$ $\times$ 4 \\
      \bottomrule
    \end{tabular}
    }
  \end{minipage}%
  \hfill 
  \begin{minipage}{0.38\textwidth} 
    \centering
    \caption*{(b) Steps Conversion for a Navigation} 
    \resizebox{0.9\textwidth}{!}{ 
    \begin{tabular}{cc}
      \toprule
      End-to-end Steps & Non-end-to-end Steps \\ 
      \midrule
      3  & 23 \\
      16 & 95 \\
      42 & 300 \\
      \bottomrule
    \end{tabular}
    }
  \end{minipage}
\end{table}

\section{Navigation Case}

Figure~\ref{fig:nav_case_all} depicts a full navigation episode in which our memory‑augmented agent searches for a chair in an unfamiliar indoor scene. The seven consecutive frames show the agent’s visual observations and planned motions from entry to target acquisition. Throughout the sequence, the agent (i) reasons about semantic priors—chairs are more likely near tables or in living/dining areas; (ii) fuses transient visual evidence with its episodic memory to avoid revisiting explored regions; and (iii) selects actions that maximise expected information gain while respecting safety constraints. The case illustrates how the proposed memory module complements on‑the‑fly perception to yield efficient, goal‑directed exploration in cluttered, real‑world layouts.

\begin{figure*}[!htbp]
  \centering
  \setlength{\tabcolsep}{0pt}
  \renewcommand{\arraystretch}{0}
  \begin{tabular}{@{}c@{}}
    \includegraphics[width=\linewidth]{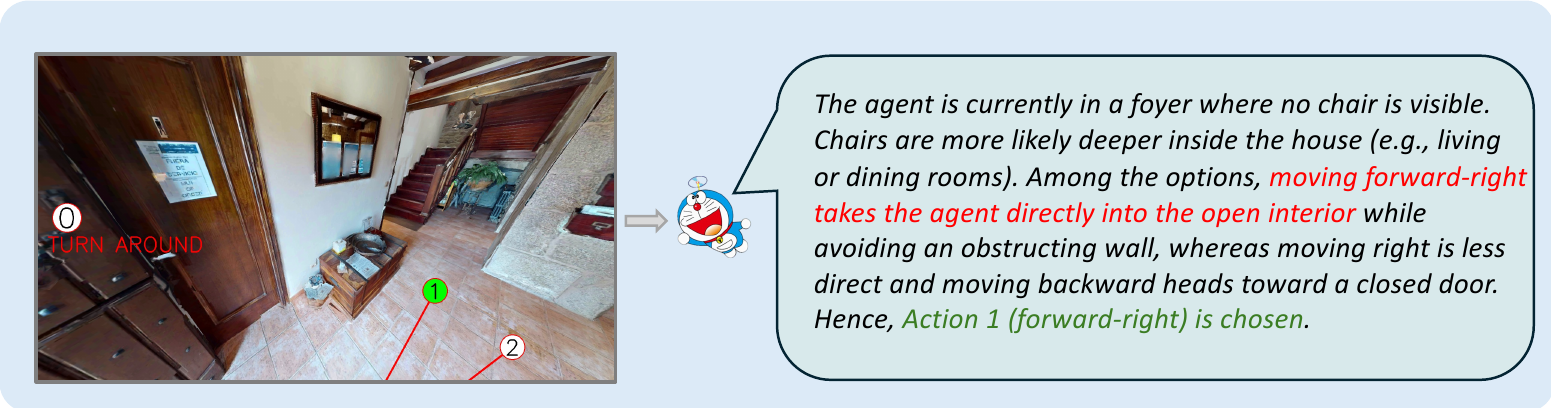}\\[3pt]
    \includegraphics[width=\linewidth]{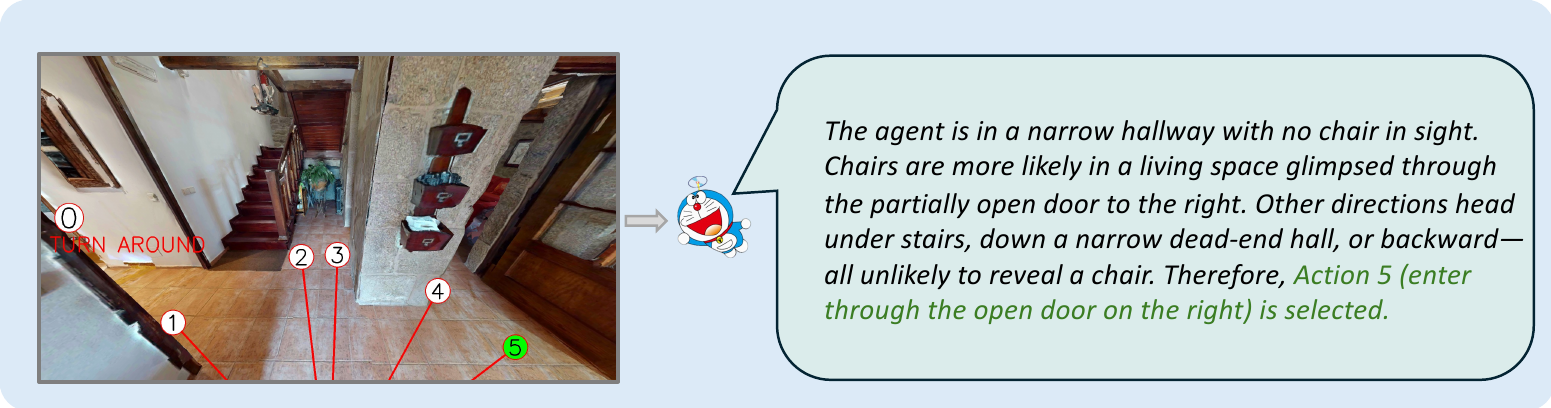}\\[3pt]
    \includegraphics[width=\linewidth]{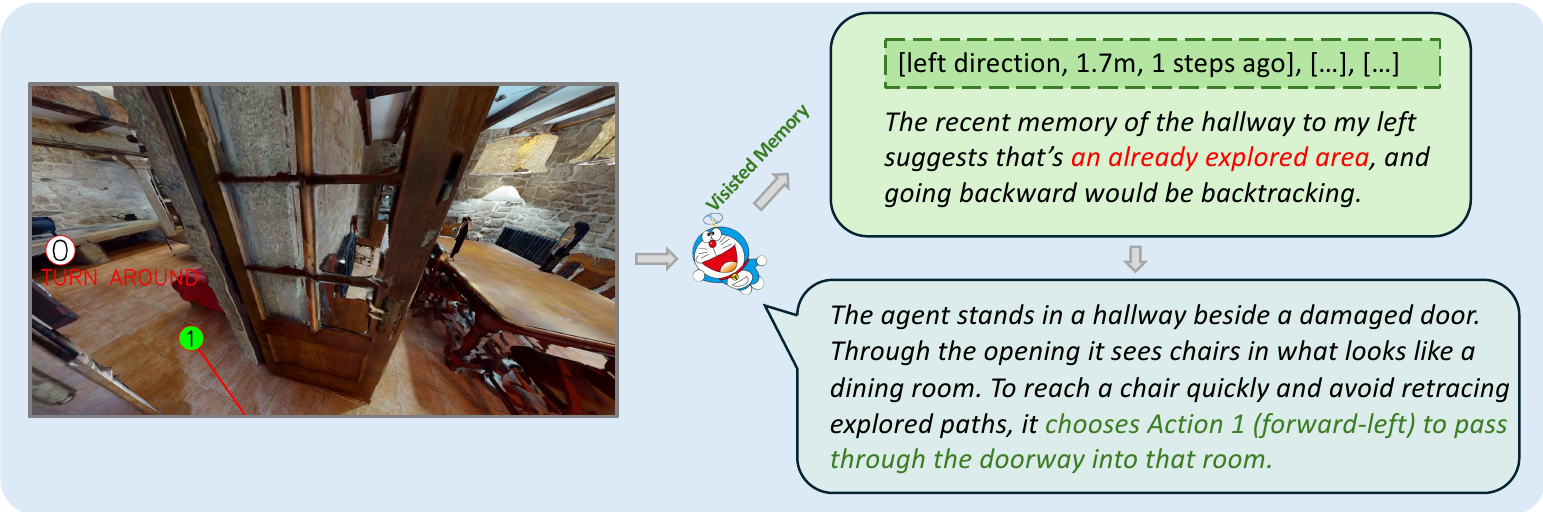}\\[3pt]
    \includegraphics[width=\linewidth]{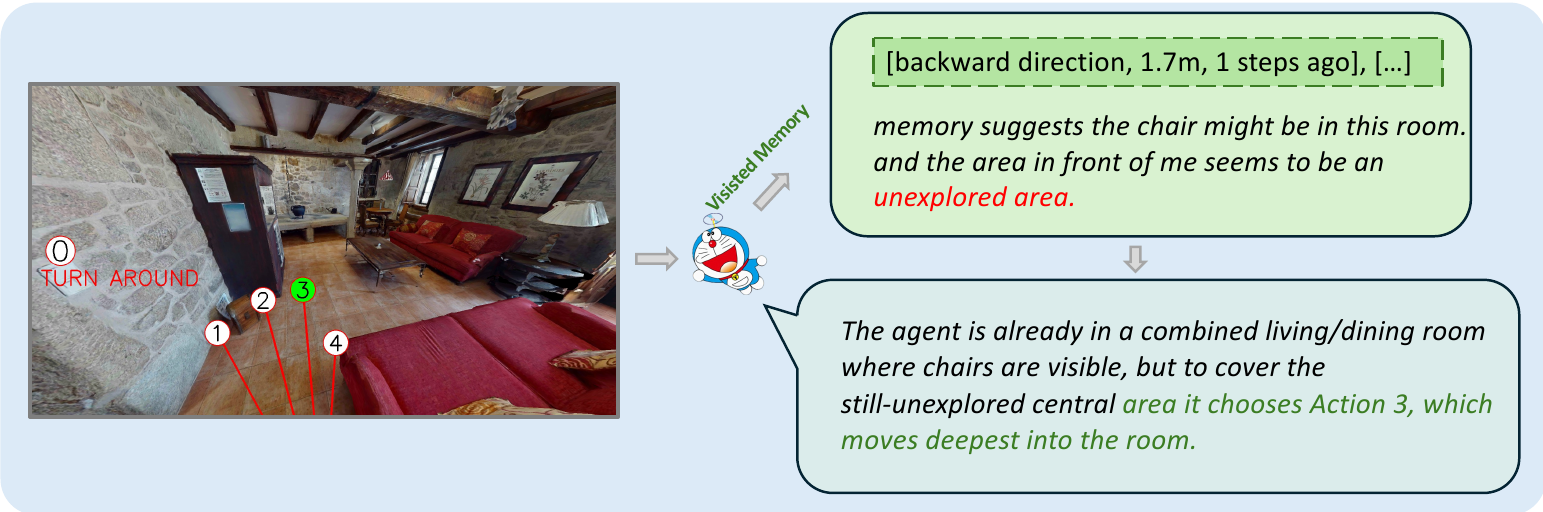}\\[3pt]
    \includegraphics[width=\linewidth]{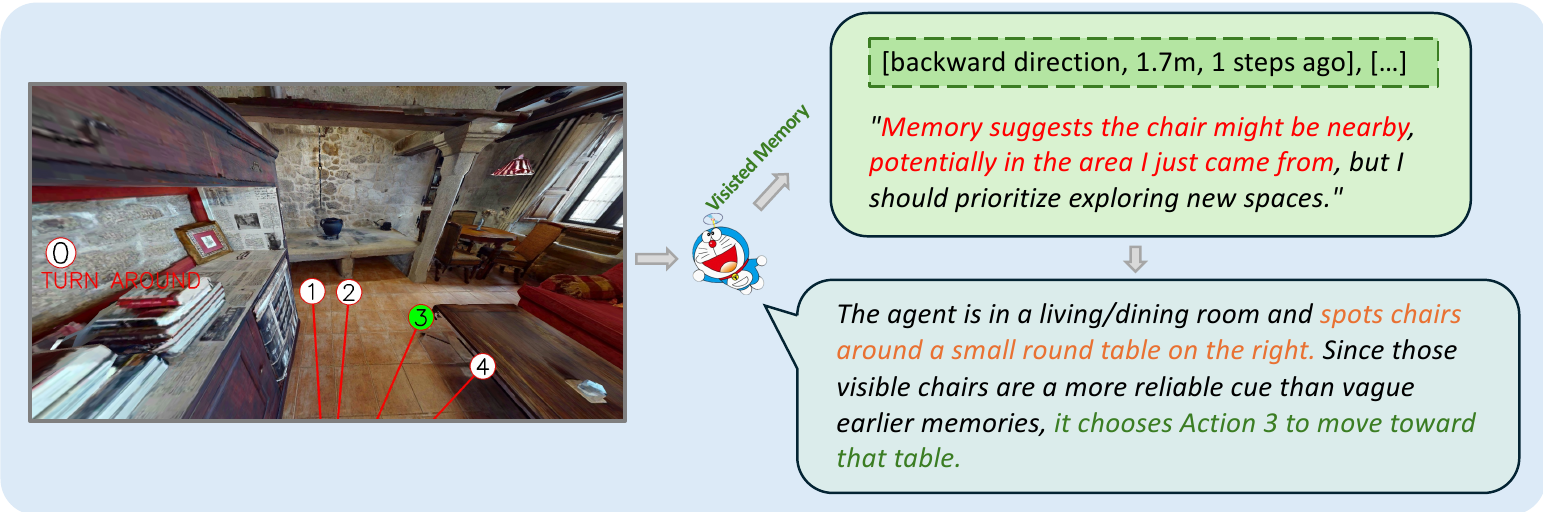}\\[3pt]
  \end{tabular}
  \captionsetup{labelformat=empty, textformat=empty, list=no} 
  \caption{} 
\end{figure*}

\begin{figure*}[!htbp]\ContinuedFloat
  \centering
  \setlength{\tabcolsep}{0pt}
  \renewcommand{\arraystretch}{0}
  \begin{tabular}{@{}c@{}}
    \includegraphics[width=\linewidth]{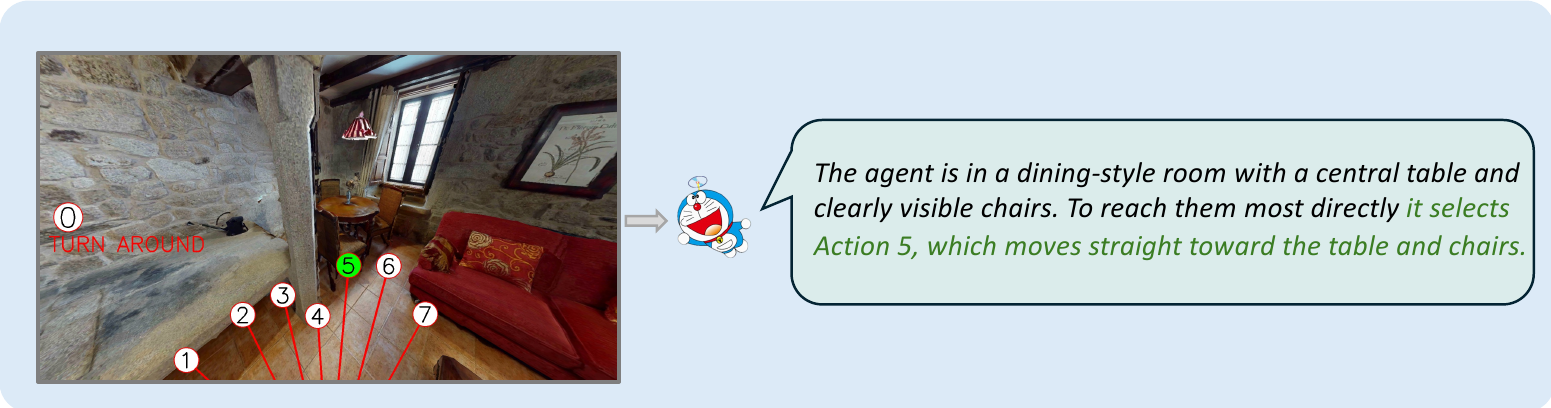}\\[3pt]
    \includegraphics[width=\linewidth]{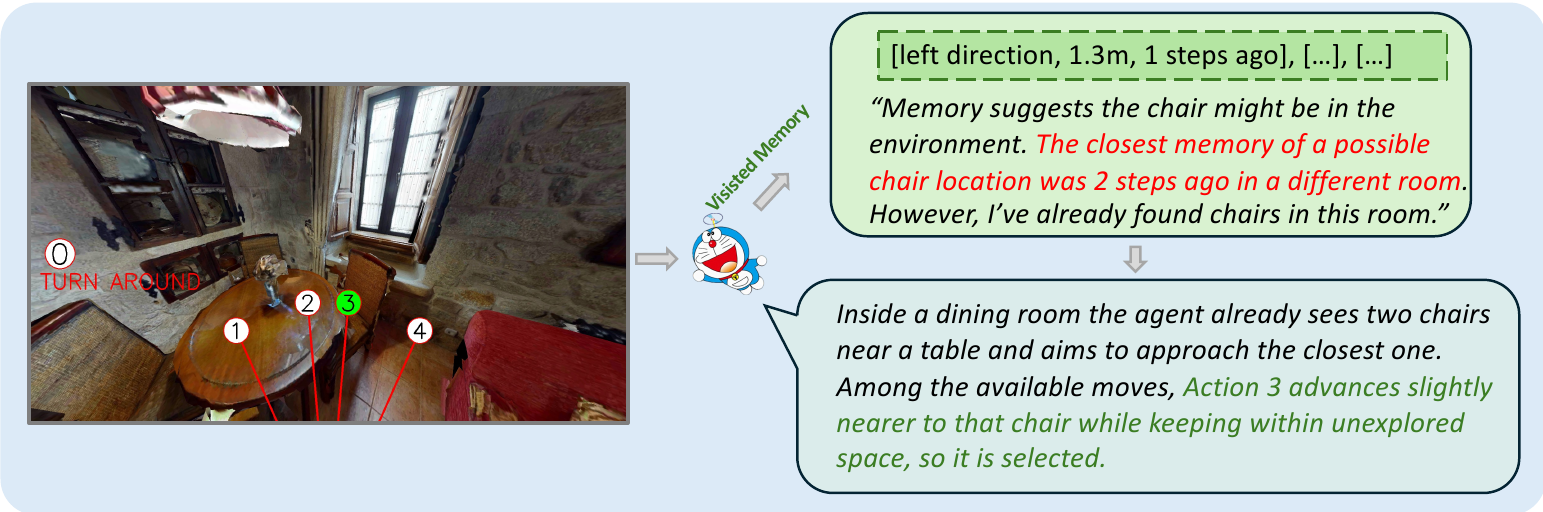}
  \end{tabular}
  \caption{%
    \textbf{Navigation case}
    Each row shows one decision step.
    \emph{Left:} the \textcolor{green!60!black}{green} circle highlights the action selected for this step.
    \emph{Upper‐right dashed green box} displays the most relevant episodic memory retrieved at this step.
    \emph{Lower‐right speech bubble} is the agent’s natural‑language rationale that fuses (i) semantic priors, (ii) current visual evidence, and (iii) memory cues.%
    }%
    \label{fig:nav_case_all}
\end{figure*}

\section{Detailed Description of AORI}\label{app:aori}

\subsection{Area Overlap Redundancy Index (AORI)}
The Area Overlap Redundancy Index (AORI) quantifies exploration efficiency through spatial overlap analysis. We formalize the computation with parameters from our implementation:

\paragraph{Parameter Basis:}
\begin{itemize}
    \item Map resolution: $5,000 \times 5,000$ grid (map\_size=5000)
    \item Voxel ray casting resolution: $60 \times 60$ (voxel\_ray\_size=60)
    \item Exploration threshold: 3 observations per voxel (explore\_threshold=3)
    \item Density scaling factor: $\eta = 0.8$ (e\_i\_scaling=0.8)
\end{itemize}

\paragraph{Step-wise Calculation:} 
For each step $t \in [1, T]$:
\begin{enumerate}
    \item Compute \textit{observed area} $A_t = \bigcup_{i=1}^t \mathcal{V}(x_i, y_i)$ where $\mathcal{V}(x,y)$ is the visible region defined by:
    \begin{equation}
        \| \mathcal{V}(x,y) \| = \frac{\text{map\_size}^2}{\text{voxel\_ray\_size}^2} \cdot \pi \quad 
    \end{equation}
    
    \item Calculate overlap ratio $r_{\text{overlap}}$:
    \begin{equation}
        r_{\text{overlap}} = \frac{\sum_{i=1}^{t-1} \mathbb{I}{[\mathcal{V}(x_t,y_t) \cap \mathcal{V}(x_i,y_i) \geq \text{explore\_threshold}]}}{t-1}
    \end{equation}
    
    \item Compute normalized density $d_{\text{normalized}}$ using Poisson expectation:
    \begin{equation}
        d_{\text{normalized}} = \min\left(1, \frac{N_{\text{obs}}}{\lambda} \right), \quad \lambda = \eta \cdot \frac{\|A_t\|}{\text{map\_size}^2} \cdot t
    \end{equation}
    where $N_{\text{obs}}$ counts voxels with $\geq$3 visits, $\lambda$ is expected active voxels
\end{enumerate}

\paragraph{Boundary Cases:}
\begin{itemize}
    \item \textit{Optimal Case} (AORI=0): $\text{When } r_{\text{overlap}} = 0 \ \& \ d_{\text{normalized}} = 0 \Rightarrow 1 - (0.8\cdot1^2 + 0.2\cdot1) = 0$
    
    \item \textit{Worst Case} (AORI=1):$\text{When } r_{\text{overlap}} = 1 \ \& \ d_{\text{normalized}} = 1 \Rightarrow 1 - (0.8\cdot0 + 0.2\cdot0) = 1$
\end{itemize}

\paragraph{Calculation Examples:}
\begin{itemize}
    \item \textit{Case1: stay still} (t=100 steps):
    \begin{equation}
        \begin{aligned}[t]
            r_{\text{overlap}} &= \frac{99}{99} = 1.0, \\
            \lambda &= 0.8 \cdot \frac{\pi (60/5000)^2}{1} \cdot 100 \approx 0.014, \\
            d_{\text{norm}} &= \min\left(1, \frac{100}{0.014}\right) = 1.0, \\
            \text{AORI} &= 1 - [0.8(1-1)^2 + 0.2(1-1)] = 1.0
        \end{aligned}
    \end{equation}
    
    \item \textit{Case2: go around} (t=500 steps):
    \begin{equation}
        \begin{aligned}[t]
            r_{\text{overlap}} &\approx \frac{38}{499} \approx 0.076, \\
            \lambda &= 0.8 \cdot \frac{\pi (60/5000)^2}{1} \cdot 500 \approx 0.069, \\
            d_{\text{norm}} &= \min\left(1, \frac{62}{0.069}\right) = 1.0, \\
            \text{AORI} &= 1 - [0.8 \times (1-0.076)^2 + 0.2 \times (1-1)] \approx 0.285
        \end{aligned}
    \end{equation}
\end{itemize}

\section{Experimental Setup Details}\label{app:setup_structured}

\paragraph{Implementation Details.} 
The maximal navigation steps per episode are set to 40. The agent's body has a radius of $ 0.17 m$ and a height of $ 1.5 m$. Its RGB-D sensors are positioned at $ 1.5 m$ height with a $-0.45$ radian downward tilt and provide a $131^\circ$ Field of View (FoV). 
For rotation, the agent selects an angular displacement corresponding to one of 60 discrete bins that uniformly discretize the $360^\circ$ range. Success requires stopping within $d_{\text{success}}=0.3$m of the target object and visually confirming it. Success requires stopping within $d_{\text{success}}=0.3 m$ of the target object and visually confirming it. Our DORAEMON framework primarily utilizes \texttt{Gemini-1.5-pro} as the VLM and \texttt{CLIP ViT-B/32} for semantic embeddings, with caching implemented for efficiency. Key hyperparameters include: topological map connection distance $\delta_{\text{connect}}=1.0$m, node update interval $S_{\text{update}}=3$ steps, $L_1$ hierarchical clustering weight $w=0.4$, AORI grid resolution $\delta_{\text{grid}}=0.1$m, minimum obstacle clearance $d_{\text{min\_obs}}=0.5$m, and various stuck detection thresholds (e.g., path inefficiency $\eta_{\text{path}} < 0.25$, small area coverage $\delta_{\text{area\_gain}} < 0.35$m$^2$, high rotation/translation ratio $\rho_{\text{rot/trans}} > 2.0$ for short paths when $\|\text{path}\| < 0.5$m) and a precision movement factor $\gamma_{\text{step}}=0.1$.

\section{Hierarchical Construction}\label{app:Hierarchical Construction}

\subsection{Level $L_3$: Observation Anchoring}
\begin{itemize}
    \item \textbf{Input}: Raw topological nodes $v_t \in \mathcal{V}$ from Eq~\ref{eq:memory_hierarchy}
    \item \textbf{Process}: Directly mapping to memory nodes
    \begin{equation}
        h_j^{(3)} = \left( \texttt{id}_j^{(3)}, L_3, \emptyset, \{v_t\} \right). 
    \end{equation}
    \item \textbf{Output}: $h_j^{(3)}$ nodes storing original $p_t, \mathbf{s}_t$ from $v_t$
\end{itemize}

\subsection{Level $L_2$: Area Formation ($L_3$ → $L_2$)}
\begin{itemize}
    \item \textbf{Input}: $h_j^{(3)}$ nodes with spatial coordinates $p_t$
    \item \textbf{Clustering}:
    \begin{enumerate}
        \item Compute combined distance:
        \begin{equation}
            d_{\text{comb}} = 0.4\|p_i-p_j\|_2 + 0.6\left(1-\frac{\mathbf{s}_i\cdot\mathbf{s}_j}{\|\mathbf{s}_i\|\|\mathbf{s}_j\|}\right). 
        \end{equation}
        \item Apply adaptive threshold:
        \begin{equation}
            \theta_1' = \begin{cases} 
                1.5\theta_1 & (|\mathcal{O}| >20) \\
                0.8\theta_1 & (|\mathcal{O}| <10) \\
                \theta_1 & \text{otherwise}. 
            \end{cases}
        \end{equation}
        \item Generate clusters using scipy.linkage + fcluster
    \end{enumerate}
    \item \textbf{Functional Labeling}:
    \begin{equation}
        \texttt{area\_type} = \arg\max_t\sum_{v\in \mathcal{C}_j^{(2)}}\sum_{k\in K_t}\mathbb{I}[k \in v.L_t]. 
    \end{equation}
    \item \textbf{Output}: $h_m^{(2)}$ nodes with:
    \begin{itemize}
        \item Parent: $h_n^{(1)}$ ($L_1$ room node).
        \item Children: $\{h_j^{(3)}\}$ ( observations).
        \item Spatial boundary: Convex hull of $p_t$ positions.
    \end{itemize}
\end{itemize}

\subsection{Level $L_1$: Room Formation ($L_2$ → $L_1$)}
\begin{itemize}
    \item \textbf{Input}: $h_m^{(2)}$ areas with spatial centroids $P_A$
    \item \textbf{Two-stage Clustering}:
    \begin{enumerate}
        \item \textit{Spatial Pre-clustering}:
        \begin{equation}
            C_{\text{spatial}} = \text{fcluster}(\text{linkage}(d_{\text{spatial}}), \theta_2=3.0\text{m}). 
        \end{equation}
        \item \textit{Functional Refinement}:
        \begin{equation}
            \mathcal{F}_s = \{\mathcal{A}_{s,f} | f=\text{MapToRoomFunction}(\texttt{area\_type})\}. 
        \end{equation}
    \end{enumerate}
    \item \textbf{Output}: $h_n^{(1)}$ nodes containing:
    \begin{itemize}
        \item Parent: $h_0^{(0)}$ ($L_0$ root)
        \item Children: $\{h_m^{(2)}\}$ ($L_2$ areas)
    \end{itemize}
\end{itemize}

\subsection{Level $L_0$: Environment Root}
\begin{itemize}
    \item \textbf{Input}: All $h_n^{(1)}$ room nodes
    \item \textbf{Consolidation}:
    \begin{equation}
        h_0^{(0)} = \left( \texttt{GLOBAL\_ROOT}, L_0, \emptyset, \{h_n^{(1)}\} \right). 
    \end{equation}
    \item \textbf{Function}: Global access point for memory queries
\end{itemize}

\section{Memory Retrieval Scoring Details}\label{app:scoring_details}
\subsection{Scoring Function Decomposition}
The retrieval score combines four evidence components through weighted summation:
\begin{equation}
S(h_i) = {0.45S_{\text{sem}} + 0.30S_{\text{spa}} + 0.20S_{\text{key}} + 0.05S_{\text{time}}}. 
\label{eq:score_components}
\end{equation}

\subsection{Component Specifications}

\subsubsection{Semantic Similarity}
\begin{itemize}
    \item \textbf{Input}: CLIP embeddings $\mathbf{s}_q$ (query) and $\mathbf{s}_i$ (node)
    \item \textbf{Calculation}:
    \begin{equation}
    S_{\text{sem}} = \frac{1}{2}\left(1 + \frac{\mathbf{s}_q^\top \mathbf{s}_i}{\|\mathbf{s}_q\| \|\mathbf{s}_i\|}\right) \in [0,1]. 
    \end{equation}
\end{itemize}

\subsubsection{Spatial Proximity}
\begin{itemize}
    \item \textbf{Input}: Agent position ${p}_a$, node position ${p}_i$
    \item \textbf{Decay function}:
    \begin{equation}
    S_{\text{spa}} = \exp\left(-\frac{\|{p}_a - {p}_i\|_2}{5.0}\right). 
    \end{equation}
\end{itemize}

\subsubsection{Keyword Relevance} 
\begin{itemize}
    \item \textbf{Input}: Query terms $T$, node keywords $K_i$ (from $L_t$)
    \item \textbf{Matching score}:
    \begin{equation}
    S_{\text{key}} = \frac{|T \cap K_i|}{\max(|T|,1)}. 
    \end{equation}
\end{itemize}

\subsubsection{Temporal Recency}
\begin{itemize}
    \item \textbf{Input}: Current time $t_c$, observation time $t_i$
    \item \textbf{Decay model}:
    \begin{equation}
    S_{\text{time}} = \exp\left(-\frac{|t_c - t_i|}{600}\right). 
    \end{equation}
\end{itemize}

\subsection{Parameter Configuration}
\begin{table}[ht]
\centering
\caption{Scoring Component Weights}
\label{tab:score_weights}
\begin{tabular}{lcc}
\toprule
Component & Symbol & Value \\
\midrule
Semantic Similarity & $\alpha_{\text{sem}}$ & 0.45 \\
Spatial Proximity & $\alpha_{\text{spa}}$ & 0.30 \\
Keyword Relevance & $\alpha_{\text{key}}$ & 0.20 \\
Temporal Recency & $\alpha_{\text{time}}$ & 0.05 \\
\bottomrule
\end{tabular}
\end{table}

\subsection{Search Process}
The beam search executes through these discrete phases:

\paragraph{Initialization Phase}
\begin{itemize}
    \item Start from root node(s): $\mathcal{F}_0 = \{h_{\text{root}}\}$
    \item Set beam width: $B=5$
\end{itemize}

\paragraph{Iterative Expansion}
For each hierarchy level $l \in \{L_3,L_2,L_1,L_0\}$:
\begin{itemize}
    \item Score all children: $S(h_{\text{child}}) \forall h_{\text{child}} \in \mathcal{C}(h_j), h_j \in \mathcal{F}_l$
    \item Select top-$B$ nodes
\end{itemize}

\paragraph{Termination Conditions}
\begin{itemize}
    \item \textbf{Success}: Reached $L_0$ nodes and selected top-$K$ results
    \item \textbf{Failure}: No nodes satisfy $S(h_i) > 0.4$ threshold
\end{itemize}

\subsection{Computational Properties}
\begin{itemize}
    \item \textbf{Time Complexity}: $O(B \cdot D)$ for depth $D=4$
    \item \textbf{Memory Complexity}: $O(B)$ nodes per level
    \item \textbf{Score Normalization}:
    \begin{equation}
    \sum_{k\in\{\text{sem,spa,key,time}\}} \alpha_k = 1.0. 
    \end{equation}
\end{itemize}

\section{Chain-of-Thought Prompt}
\label{app:cot_prompt}

Our Exec-VLM leverages a structured Chain-of-Thought (CoT) prompt to guide the decision-making process. The complete prompt is provided below:

\begin{lstlisting}
TASK: NAVIGATE TO THE NEAREST [TARGET_OBJECT], and get as close to it as possible. 
Use your prior knowledge about where items are typically located within a home. 
There are [N] red arrows superimposed onto your observation, which represent potential actions. 
These are labeled with a number in a white circle, which represent the location you would move to if you took that action. 
[TURN_INSTRUCTION]

Let's solve this navigation task step by step:

1. Current State Analysis: What do you observe in the environment? What objects and pathways are visible? 
   Look carefully for the target object, even if it's partially visible or at a distance.

2. Memory Integration: Review the memory context below for clues about target location.
   - Pay special attention to memories containing or near the target object
   - Use recent memories (fewer steps ago) over older ones
   - Consider action recommendations based on memory
   
3. Goal Analysis: Based on the target and home layout knowledge, where is the [TARGET_OBJECT] likely to be?

4. Scene Assessment: Quickly evaluate if [TARGET_OBJECT] could reasonably exist in this type of space:
   - If you're in an obviously incompatible room (e.g., looking for a [TARGET_OBJECT] but in a clearly different room type), choose action 0 to TURN AROUND immediately

5. Path Planning: What's the most promising direction to reach the target? Avoid revisiting 
   previously explored areas unless necessary. Consider:
   - Available paths and typical room layouts
   - Areas you haven't explored yet

6. Action Decision: Which numbered arrow best serves your plan? Return your choice as {"action": <action_key>}. Note:
   - You CANNOT GO THROUGH CLOSED DOORS, It doesn't make any sense to go near a closed door.
   - You CANNOT GO THROUGH WINDOWS AND MIRRORS
   - You DO NOT NEED TO GO UP OR DOWN STAIRS
   - Please try to avoid actions that will lead you to a dead end to avoid affecting subsequent actions, unless the dead end is very close to the [TARGET_OBJECT]
   - If you see the target object, even partially, choose the action that gets you closest to it
\end{lstlisting}

\section{Detailed Description of Baseline}\label{app:basline}
To assess the performance of \emph{DORAEMON}, we compare it with \textbf{16} recent baselines for (zero‑shot) object‑goal navigation. Summaries are given below.

\textbf{ProcTHOR}~\cite{deitke2022}: A procedurally–generated 10K‑scene suite for large‑scale Embodied AI.  

\textbf{ProcTHOR\_ZS}~\cite{deitke2022}: ProcTHOR\_ZS trains in ProcTHOR and evaluates zero‑shot on unseen iTHOR/RoboTHOR scenes to test cross‑domain generalisation.

\textbf{SemEXP}~\cite{chaplot2020object}: Builds an online semantic map and uses goal‑oriented exploration to locate the target object efficiently, achieving state‑of‑the‑art results in Habitat ObjectNav 2020.

\textbf{Habitat‑Web}~\cite{ramrakhya2022habitat}: Collects large‑scale human demonstrations via a browser interface and leverages behaviour cloning to learn object‑search strategies.

\textbf{PONI}~\cite{ramakrishnan2022poni}: Learns a potential‑field predictor from static supervision, enabling interaction‑free training while preserving high navigation success.

\textbf{ZSON}~\cite{majumdar2022zson}: Encodes multimodal goal embeddings (text + images) to achieve zero‑shot navigation towards previously unseen object categories.

\textbf{PSL}~\cite{sun2024prioritized}: Prioritised Semantic Learning selects informative targets during training and uses semantic expansion at inference for zero‑shot instance navigation.

\textbf{Pixel‑Nav}~\cite{cai2024bridging}: Introduces pixel‑guided navigation skills that bridge foundation models and ObjectNav, relying solely on RGB inputs.

\textbf{SGM}~\cite{zhang2024imagine}: “Imagine Before Go” constructs a self‑supervised generative map to predict unseen areas and improve exploration efficiency.

\textbf{ImagineNav}~\cite{zhao2024imaginenav}: Prompts vision–language models to imagine future observations, guiding the agent toward information‑rich viewpoints.

\textbf{CoW}~\cite{gadre2023cows}: Establishes the “Cows on Pasture” benchmark for language‑driven zero‑shot ObjectNav and releases baseline policies without in‑domain training.

\textbf{ESC}~\cite{zhou2023esc}: Employs soft commonsense constraints derived from language models to bias exploration, markedly improving zero‑shot success over CoW.

\textbf{L3MVN}~\cite{Yu_2023}: Utilises large language models to reason about likely room sequences, while a visual policy executes the suggested path.

\textbf{VLFM}~\cite{yokoyama2024vlfm}: Combines VLM goal‑localisation with frontier‑based exploration, removing the need for reinforcement learning or task‑specific fine‑tuning.

\textbf{VoroNav}~\cite{wu2024voronav}: Simplifies the search space via Voronoi partitions and pairs this with LLM‑driven semantic planning for improved zero‑shot performance.

\textbf{TopV‑Nav}~\cite{zhong2024topv}: Lets a multimodal LLM perform spatial reasoning directly on top‑view maps, with adaptive visual prompts for global–local coordination.

\textbf{SG‑Nav}~\cite{yin2024sg}: Online builds a 3D scene graph and uses hierarchical Chain‑-of-‑Thought prompting so an LLM can infer probable target locations.

\end{document}